\documentclass{cup-journal}

\makeatletter
\patch@numeric@authors
\makeatother

\makeatletter
\def\cup@journal@name{}
\makeatother

\usepackage{xpatch}
\makeatletter
\xpatchcmd{\@maketitle}
  {\parbox[b]{\dimexpr\textwidth-26mm\relax}{%
    \firstheadsize{\journalnamefont\cup@journal@name} 
    (\cup@year), {\volumefont\cup@vol}, \thepage--\pageref{LastPage}
    \par%
  }\hfill\par
  \vspace*{\baselineskip}}
  {}{}{}
\makeatother

\makeatletter
\xpatchcmd{\@maketitle}
  {{\sffamily\bfseries\color{structure@color}%
    \MakeUppercase{\convertchar{\cup@manuscript}{-}{ }}\par}}
  {}{}{}
\makeatother

\colorlet{tbrowcolor}{white}
\colorlet{tbheadcolor}{white}
\usepackage{amsmath, amssymb, amsthm, mathtools}

\usepackage{graphicx}
\usepackage{booktabs}
\usepackage{caption}
\usepackage{subcaption}
\usepackage{siunitx}
\usepackage{float}
\usepackage[frozencache,cachedir=.]{minted}
\usepackage{xcolor}

\usepackage[hidelinks]{hyperref}
\usepackage[nameinlink,noabbrev]{cleveref}

\title{Geospatial Machine Learning Libraries}

\author{Adam J. Stewart}
\affiliation{Chair of Data Science in Earth Observation, Technical University of Munich}
\alsoaffiliation{Munich Center for Machine Learning}

\author{Caleb Robinson}
\affiliation{AI for Good Research Lab, Microsoft}

\author{Arindam Banerjee}
\affiliation{Siebel School of Computing and Data Science, University of Illinois Urbana-Champaign}


\begin{document}
\maketitle

\begin{abstract}
Recent advances in machine learning have been supported by the emergence of domain-specific software libraries, enabling streamlined workflows and increased reproducibility. For geospatial machine learning (GeoML), the availability of Earth observation data has outpaced the development of domain libraries to handle its unique challenges, such as varying spatial resolutions, spectral properties, temporal cadence, data coverage, coordinate systems, and file formats. This chapter presents a comprehensive overview of GeoML libraries, analyzing their evolution, core functionalities, and the current ecosystem. It also introduces popular GeoML libraries such as TorchGeo, eo-learn, and Raster Vision, detailing their architecture, supported data types, and integration with ML frameworks. Additionally, it discusses common methodologies for data preprocessing, spatial--temporal joins, benchmarking, and the use of pretrained models. Through a case study in crop type mapping, it demonstrates practical applications of these tools. Best practices in software design, licensing, and testing are highlighted, along with open challenges and future directions, particularly the rise of foundation models and the need for governance in open-source geospatial software. Our aim is to guide practitioners, developers, and researchers in navigating and contributing to the rapidly evolving GeoML landscape.
\end{abstract}

\section{Introduction}\label{sec:Intro}
Machine learning (ML) software libraries have played a foundational role in both shaping ML research and the commoditization of ML models. Libraries such as \textbf{scikit-learn} \citep{sklearn}, \textbf{TensorFlow} \citep{tensorflow}, and \textbf{PyTorch} \citep{pytorch} incorporate best practices, provide composable APIs, and abstract low-level engineering complexity, allowing researchers and practitioners to focus on experimentation, implementation of higher-level features, and conceptual development. A well designed software library provides a platform that users can build upon to pursue their agendas. As a result, the accessibility and maturity of ML tooling are tightly coupled with research progress.

Beyond general-purpose frameworks, domain-specific libraries further lower the barriers to research in particular subfields. These libraries often provide data loaders, evaluation metrics, pretrained models, and architecture templates specific to that domain. For example, the \textbf{torchvision} library \citep{marcel2010torchvision} simplifies experimentation in computer vision by including standard datasets (e.g., CIFAR \citep{krizhevsky2009learning}, ImageNet \citep{deng2009imagenet}), data augmentation routines, and reference implementations of neural network architectures. In natural language processing (NLP), the \textbf{Hugging Face} ecosystem has standardized tokenization, model configuration, and inference across transformer-based architectures \citep{wolf2019huggingface}, enabling large-scale evaluation and pretraining. Similar domain-specific tooling ecosystems exist in reinforcement learning (e.g., \textbf{OpenAI Gym} \citep{brockman2016openai}), speech (e.g., \textbf{torchaudio} \citep{yang2022torchaudio}), and time series (e.g., \textbf{GluonTS} \citep{alexandrov2020gluonts}), all of which have substantially influenced research agendas and benchmarks.

In contrast, the development of libraries tailored to geospatial machine learning (GeoML) has lagged behind. This gap is notable given the increasing volume and accessibility of satellite and aerial imagery from public and commercial platforms. Remotely-sensed data support a wide range of applications --- including land cover mapping \citep{karra2021global}, agricultural monitoring \citep{MULLA2013358}, disaster response \citep{van2013remote}, and climate modeling \citep{rolnick2019tackling} --- but present a set of challenges not typically encountered in standard ML workflows. As presented in the introduction chapter of this book, Earth observation images vary in spatial resolution (from millimeters to tens of kilometers per pixel), spectral characteristics (e.g., 3-band RGB imagery vs.\ 13-band multispectral Sentinel-2 imagery vs.\ 242-band Hyperion imagery), and temporal cadence (every five minutes from the NASA GOES satellites, to on-demand tasking for commercial high-resolution satellites, to every 16 days for Landsat 9). Further complications include non-uniform data coverage over the Earth, cloud occlusion, differing coordinate reference systems (CRS), and file formats optimized for spatial indexing rather than machine learning \citep{rolf2024position}. These characteristics require preprocessing steps such as reprojection, resampling, and cloud masking, none of which are directly supported by conventional ML data pipelines. As a result, integrating geospatial data into an ML workflow typically requires manual interfacing with specialized libraries such as \textbf{GDAL} \citep{gdal2022}, creating substantial friction between data access and model development.

Despite these challenges, recent years have seen progress toward standardizing workflows for GeoML through the development of libraries explicitly designed for geospatial data in machine learning contexts. \textbf{TorchGeo} \citep{torchgeo}, for example, builds on PyTorch to provide dataset abstractions, spatial sampling utilities, and support for georeferenced imagery and vector data. Other libraries, such as \textbf{eo-learn} \citep{eolearn} and \textbf{Raster Vision} \citep{rastervision}, offer higher-level frameworks for remote sensing tasks, including land cover classification, semantic segmentation, and change detection. These libraries attempt to bridge the gap between geospatial data infrastructure and modern ML tooling, offering primitives for common operations such as patch extraction, label alignment, and tiling. However, the ecosystem remains fragmented, with limited support for tasks such as temporal modeling, sensor fusion, or distributed inference over petabyte-scale datasets. Moreover, unlike domains such as vision and language modeling, GeoML currently lacks a unifying toolkit that combines data curation, model training, and deployment into an integrated framework, though considerable progress has been made in recent years. While the current chapter primarily focuses on Python libraries as much of the active development, GeoML developments in other languages such as R have similar or related issues.

\section{Methodology}\label{sec:Method}

By definition, GeoML libraries must support two core tasks: processing geospatial data, and integrating it into machine learning workflows. Their functionality spans input/output (I/O) operations, internal data representations, and dataset handling. 

Most GeoML libraries rely on the Geospatial Data Abstraction Library (GDAL) either directly or indirectly to support I/O operations. GDAL provides a consistent API for interacting with a wide range of raster and vector data formats --- as of September 2025, it includes 155 raster drivers and 83 vector drivers, covering formats such as GeoTIFF, Zarr, HDF5, NetCDF, Esri Shapefile, and GeoPackage. In Python, libraries like \textbf{rasterio} \citep{rasterio} and \textbf{fiona} \citep{Gillies_Fiona_2024} wrap GDAL's API and provide an additional layer of abstraction and additional features (e.g., vectorization, rasterization, and fine-grained management of environment variables to control GDAL behavior). In R, \textbf{terra} \citep{terra} and \textbf{sf} \citep{RJ-2018-009} provide similar support for raster and vector operations at a higher level of abstraction. Even higher-levels of abstraction exist through libraries such as \textbf{xarray} \citep{xarray}, \textbf{rioxarray} \citep{rioxarray}, and \textbf{geopandas} \citep{geopandas}, which provide array- or dataframe-based abstractions that retain coordinate reference system (CRS) metadata and spatial transforms. Some geospatial file formats can also be accessed through alternative backends. \textbf{h5py} \citep{h5py}, \textbf{zarr} \citep{zarr}, and \textbf{tifffile} \citep{tifffile} all provide direct interfaces for HDF5, Zarr, and TIFF files, respectively. Similarly, GeoJSON files can be read using any standard JSON parsers, although geospatial-aware libraries (like fiona) ensure proper CRS interpretation and coordinate handling. A key feature throughout all GeoML libraries is how they ensure that the geospatial metadata of inputs are taken into account --- any corresponding outputs will need this information in order to be georeferenced for downstream analysis. Machine learning libraries that do not support this require users to (a) have knowledge of how geospatial metadata works and (b) develop ad-hoc methods for propagating geospatial metadata, both of which can easily allow for bugs to be introduced. 

Another key feature of GeoML libraries is how spatial and temporal data are represented within the library. Spatial and temporal joins, in particular, are common operations for setting up GeoML experiments. For example, training a model to predict whether land is cropland or not from a time series of satellite imagery observations over the planting season of that area requires several key steps. First, representing a time series of multispectral satellite imagery over the same location requires a temporal join with corresponding decisions. Does the library itself allow for the temporal join to be performed? How will missing observations be handled and how will the time dimension be represented (densely in 4D, sparsely with pointers to 3D tensors)? How are layers with differing coordinate reference systems and spatial resolutions handled? Second, a spatial join between the resulting imagery stack and a known cropland mask is necessary in order to create labeled examples that can be fed into a machine learning pipeline. If there are many different cropland masks, or satellite imagery layers, then a spatial index will be needed to ensure that intersections between the layers can be computed quickly.  Further, the spatial join must also take different coordinate reference systems and/or spatial resolutions into account --- some layers will likely need to be reprojected and resampled in an appropriate manner before further analysis can be done. The cropland masks could also be in a vector format and need to be rasterized or otherwise transformed into a format that can be used directly in modeling.

GeoML libraries also differ in how they support \textit{benchmark} datasets. The GeoML community has proposed a growing set of benchmark datasets intended to facilitate algorithmic comparison and standardized evaluation on a wide variety of tasks. These datasets typically pair remotely-sensed imagery with structured labels, such as land cover classifications \citep{robinson2019large}, object boundaries \citep{kerner2025fields}, or change detection masks \citep{chen2020spatial}. Unlike domains such as computer vision or NLP, geospatial benchmark datasets often lack standardized data loaders or preprocessing routines. 
As a result, researchers are frequently required to write custom scripts for parsing, tiling, and aligning data. This introduces inconsistencies and hinders reproducibility. GeoML libraries can substantially reduce this friction by implementing standardized, reusable data loaders for popular benchmarks. Features that distinguish high-quality dataset support include:
\begin{itemize}
    \item adherence to published train/validation/test splits;
    \item support for user-defined or random splits, including spatial stratification;
    \item compatibility with PyTorch or TensorFlow \texttt{Dataset} abstractions, enabling seamless integration into training pipelines;
    \item additional utilities such as on-the-fly reprojection, resampling, normalization, augmentation, or class remapping; and
    \item built-in visualization tools for inspecting inputs, labels, and predictions.
\end{itemize}

Taken together, these capabilities define the core functionality and value of a GeoML library. By abstracting the geospatial data pipeline from raw file access and metadata handling to dataset construction and integration with ML training loops, these libraries allow researchers and practitioners to focus on model development rather than infrastructure. However, the design choices made at each stage, such as metadata preservation, spatial joining logic, temporal indexing strategies, and dataset standardization, directly affect model reproducibility, comparability, and deployment. In the remainder of the chapter we highlight the history of GeoML libraries, break down existing ``state-of-the-art'' libraries along the dimensions we describe here, expand on our example of cropland mapping, and finally discuss best practices for GeoML library development from a software engineering standpoint.

\section{State of the Art and Current Developments}\label{sec:SOTA}

Before jumping in to the current state of GeoML library development, it helps to first understand how we got there.

\subsection{A Brief History}

The development of GeoML software libraries has evolved alongside the rise of machine learning in computer vision. Beginning in the early-2000s, researchers began repurposing general-purpose machine learning frameworks to work with the unique formats and challenges of satellite and aerial imagery. Over time, this ad hoc adaptation gave way to a more intentional, ecosystem-wide effort to build specialized tools for Earth observation data.

\begin{figure}[tbh!]
    \centering
    \includegraphics[width=0.8\linewidth]{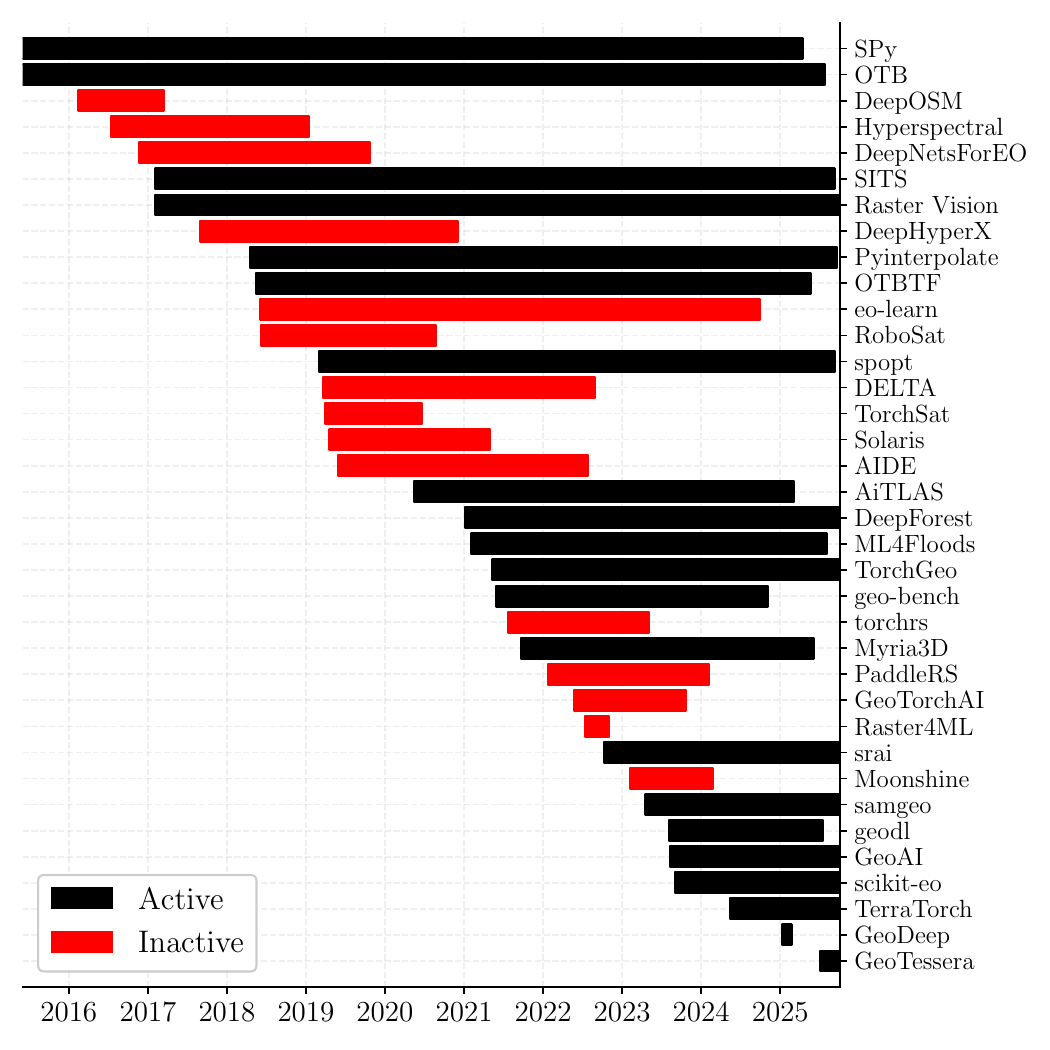}
    \caption{Timeline of GeoML library development. Libraries that are under active development (defined as a new commit within the last year) are shown in black. Inactive libraries are shown in red. SPy and OTB development stretches back to 2001 and 2006, respectively, and are truncated to focus on more recent developments. Bars denote earliest and most recent commit (as of September 2025).}
    \label{fig:timeline}
\end{figure}

One of the earliest known GeoML libraries, \textbf{SPy} (Spectral Python), began development in 2001 with simple iterative clustering methods for hyperspectral imagery, later adding $k$-means clustering and perceptron classifiers \citep{spy}. The library predates most modern ML frameworks, and all methods are implemented from scratch in NumPy \citep{2020NumPy-Array}. SPy has successfully survived several transitions (Python 2 $\rightarrow$ 3, CVS $\rightarrow$ SVN $\rightarrow$ Git) and is still maintained to this day. \textbf{OTB} (Orfeo ToolBox) is another popular software library for processing optical, multispectral, and radar imagery \citep{grizonnet2017orfeo}. Initial development began in 2006 by the French national space agency (CNES) as part of the Orfeo Program, targeting high resolution imagery from the Orfeo constellation (Pléiades and COSMO-Skymed). OTB offered early support for support vector machines (SVMs) via LibSVM \citep{chang2011libsvm} and later added support for gradient boosted decision trees, $k$-nearest neighbors ($k$-NNs), and multilayer perceptrons (MLPs) from OpenCV \citep{bradski2000opencv} and $k$-means and random forests from the Shark machine learning library \citep{shark}. OTB itself is written in C++, with a command-line interface, Python bindings, and an official QGIS plugin.

The earliest known geospatial deep learning libraries began development in 2016. TrailBehind introduced \textbf{DeepOSM} \citep{deeposm}, which built on top of the now-defunct TFLearn \citep{tflearn} and supported automatic downloading and preprocessing of labeled imagery for road network detection using OpenStreetMap (OSM) \citep{openstreetmap}. \textbf{Hyperspectral} was developed by researchers at IIT Kharagpur, with early TensorFlow-based experiments with MLPs and CNNs on hyperspectral imagery \citep{hyperspectral}. Later that year, ONERA and IRISA released \textbf{DeepNetsForEO} \citep{deepnetsforeo}, initially built on Caffe \citep{caffe} and later ported to PyTorch. Notably, it was the first library to offer pretrained models on multispectral remote sensing imagery. All three libraries were discontinued within a few years of release. They were never packaged for PyPI or other standard software registries, and instead required knowledge of Git and Docker for installation. Nonetheless, they established foundational ideas --- especially the importance of pipelines for data handling --- that would influence the next generation of GeoML libraries.

In 2017, the first broadly adopted, formally maintained geospatial deep learning libraries were introduced. In the R ecosystem, \textbf{SITS} (Satellite Image Time Series) was launched \citep{sits}. SITS is a library developed by e-sensing with a focus on time series analysis and data cube construction from providers such as AWS, the Microsoft Planetary Computer, and Copernicus Data Space. Built on R bindings to libtorch, SITS emphasizes accessibility, with formal documentation, CRAN (and more recently PyPI) packaging, and a free online book \citep{sitsbook}. One day after the initial commit of SITS, in the Python ecosystem, Azavea started work on \textbf{Raster Vision} \citep{rastervision}. Raster Vision is a modular framework for training and deploying deep learning models on arbitrary raster datasets. Initially built on TensorFlow, it migrated to PyTorch in 2019. The authors of DeepNetsForEO returned with \textbf{DeepHyperX}, a library for hyperspectral classification using scikit-learn and PyTorch \citep{deephyperx}. However, its development fractured across multiple forks on GitLab and GitHub and was later abandoned.

Four influential libraries followed in 2018. \textbf{Pyinterpolate} was introduced by Dataverse Labs, with support for several geostatistical interpolation techniques and learning-based Kriging methods \citep{pyinterpolate}. \textbf{OTBTF} was created as a TensorFlow and Keras plugin for Orfeo ToolBox and provides both Python and C++ APIs, but is only distributed through Docker due to its compilation requirements \citep{otbtf}. Sentinel Hub released \textbf{eo-learn}, a scikit-learn-compatible pipeline that offers strong integration with Sentinel Hub products and showed the possibility of cloud native workflows \citep{eolearn}. Mapbox released \textbf{RoboSat}, built on PyTorch and torchvision, which offered a command-line interface for data preparation, training, and postprocessing \citep{robosat}. RoboSat was eventually discontinued and briefly revived as RoboSat.pink, then Neat-EO.pink, before being abandoned again in 2020.

In 2019, the ecosystem diversified with several new projects. The Python Spatial Analysis Library (PySAL) introduced \textbf{spopt}, providing several spatial optimization techniques including clustering methods from scikit-learn \citep{spopt2022}. NASA released \textbf{DELTA}, a TensorFlow-based toolkit for large-image tiling and semantic segmentation \citep{delta}. The \textbf{TorchSat} project attempted to adapt torchvision to the geospatial domain \citep{torchsat}. CosmiQ Works launched \textbf{Solaris}, a PyTorch framework for object detection \citep{solaris}. Finally, Microsoft's AI for Earth team created \textbf{AIDE} \citep{aide}, a modular web-based annotation interface with Detectron2 \citep{wu2019detectron2} integration. While each offered novel features, none of these libraries besides spopt were maintained beyond 2022.

The COVID-19 pandemic brought a slowdown in 2020. Only one major library, \textbf{AiTLAS}, was released that year \citep{aitlas}. Developed by Bias Variance Labs, it built on PyTorch, scikit-learn, and eo-learn to support common tasks like classification and semantic segmentation using curated datasets. It also attempted the first large-scale benchmarking of more than 500 deep learning models on 22 EO classification datasets with its AiTLAS Benchmark Arena \citep{aitlasarena}.

In 2021, GeoML software development entered a new phase of stability and community-driven growth. The Weecology group at the University of Florida released \textbf{DeepForest}, targeting object detection of tree crowns in RGB drone imagery with pretrained torchvision models \citep{deepforest}. Trillium Technologies launched \textbf{ML4Floods}, an end-to-end PyTorch pipeline for flood extent estimation \citep{mateo-garcia_towards_2021}. Microsoft's AI for Good Lab and the University of Illinois Urbana-Champaign introduced \textbf{TorchGeo}, a PyTorch-native library combining the reproducibility of torchvision-style datasets with raster/vector data handling capabilities similar to Raster Vision \citep{torchgeo}. A similar project, \textbf{torchrs} was launched in parallel and was subsequently merged into TorchGeo \citep{torchrs}. Other major 2021 releases included ServiceNow's \textbf{GEO-Bench}, a foundation model benchmarking suite with 20 built-in datasets (many curated using TorchGeo) \citep{geobench}, and IGNF's \textbf{Myria3D}, uniquely focused on aerial LiDAR segmentation \citep{myria3d}.

By contrast, most libraries launched in 2022 were eventually abandoned. Baidu's \textbf{PaddleRS} \citep{paddlers} built on the PaddlePaddle deep learning framework \citep{paddlepaddle}, making it the first Chinese-developed GeoML library to gain traction. Wherobots released GeoTorch, later renamed \textbf{GeoTorchAI} \citep{geotorch}, offering scalable spatial indexing with Apache Sedona \citep{sedona}, but the project was discontinued in favor of TorchGeo. \textbf{Raster4ML}, designed for agricultural research, provided over 350 spectral indices and scikit-learn compatibility \citep{raster4ml}. Finally, Kraina AI released \textbf{srai} (Spatial Representations for Artificial Intelligence), a library providing pre-computed embeddings and spatial data visualization and processing tools \citep{srai}. Among these libraries, only srai is still maintained to this day.

A new wave of tools emerged in 2023. \textbf{Moonshine} focused on hosting pretrained semantic segmentation models but was abandoned \citep{moonshine}. Open Geospatial Solutions launched both \textbf{samgeo} \citep{samgeo}, a GUI wrapper for Meta's Segment Anything Model (SAM) \citep{sam}, and \textbf{GeoAI} \citep{geoai}, a general-purpose toolkit for satellite, LiDAR, and vector data built atop TorchGeo and scikit-learn. \textbf{Geodl}, another R library built on top of the torch and terra packages, brought 2D convolutional neural networks to the R GeoML ecosystem for the first time \citep{geodl}. \textbf{Scikit-eo} also debuted, offering another scikit-learn interface for EO tasks \citep{scikiteo}.

In 2024, IBM released \textbf{TerraTorch}, a toolkit for fine-tuning foundation models, built on TorchGeo \citep{terratorch}. Together with GEO-Bench and GeoAI, TerraTorch represents a second generation of GeoML libraries —-- those that complement instead of compete with core infrastructure by extending existing GeoML libraries like TorchGeo.

Launched in 2025, UAV4GEO's \textbf{GeoDeep} is focused on providing a lightweight solution for object detection and semantic segmentation, with dependencies only on onnxruntime and rasterio \citep{geodeep}. This library was inspired by and uses some code from Deepness \citep{deepness} and DeepForest. Cambridge's \textbf{GeoTessera} is the first library to focus solely on embeddings, providing easy access to learned representations from the Tessera foundation model~\citep{feng2025tesseratemporalembeddingssurface}.

The evolution of GeoML software has mirrored broader trends in machine learning and open-source development. While many early projects were short-lived, they introduced core abstractions --- data tiling, preprocessing pipelines, pretrained weights --- that persist in modern libraries. In recent years, the field has converged around modular, PyTorch-based tooling with strong community support, clearer maintenance practices, and increasing interoperability.

\subsection{Current Landscape}

\begin{table}[tb]
\caption{Features available in popular GeoML libraries (as of September 2025). While the majority of libraries provide general purpose data loaders and models, only a handful provide curated dataset-specific data loaders or pre-trained model weights.}
\label{tab:features}
\resizebox{1.0\textwidth}{!}{%
\begin{tabular}{@{}l>{\centering\arraybackslash}p{0.14\linewidth}rrccccc@{}}
\toprule
 & & \multicolumn{2}{c}{Feature Count} & \multicolumn{5}{c}{Feature Support} \\
\cmidrule(lr){3-4} \cmidrule(lr){5-9}
Library & ML Backend & Datasets & Weights & CLI & GUI & Reprojection & STAC & Time Series \\ \midrule
TorchGeo & PyTorch & 127 & 120 & yes & no & yes & no & partial \\
OTB & LibSVM, OpenCV, Shark & 0 & 0 & yes & yes & yes & no & no \\
TerraTorch & PyTorch & 27 & 13 & yes & no & yes & no & partial \\
DeepForest & PyTorch, TensorFlow* & 0 & 4 & no & no & no & no & no \\
Raster Vision & PyTorch, TensorFlow* & 0 & 6 & yes & no & yes & yes & partial \\
samgeo & PyTorch & 0 & 0 & no & yes & yes & no & no \\
spopt & scikit-learn & 0 & 0 & no & no & yes & no & no \\
SITS & R Torch & 22 & 0 & no & no & yes & yes & yes \\
SPy & numpy & 3 & 0 & no & no & no & no & no \\
srai & PyTorch & 0 & 0 & no & no & no & no & no \\
ML4Floods & PyTorch & 0 & 0 & no & no & yes & no & partial \\
GEO-Bench & PyTorch & 12 & 0 & no & no & no & no & no \\
scikit-eo & scikit-learn, TensorFlow & 0 & 0 & no & no & no & no & partial \\
GeoAI & PyTorch & 0 & 6 & no & no & yes & yes & no \\
Myria3D & PyTorch & 0 & 0 & partial & no & no & no & no \\
GeoTessera & scikit-learn & 0 & 0 & yes & no & yes & no & no \\
OTBTF & TensorFlow & 0 & 0 & yes & no & yes & no & no \\
GeoDeep & ONNX & 0 & 7 & yes & no & yes & no & no \\
\bottomrule
\end{tabular}
}
\footnotesize{*Support was dropped in newer releases.}
\end{table}

\begin{table}[tb]
\caption{GitHub engagement statistics for popular GeoML libraries (as of September 2025). All statistics are reported by the GitHub API. Test coverage is manually calculated when not available via Codecov and may be higher than reported in the case of failing tests.}
\label{tab:github}
\resizebox{1.0\textwidth}{!}{%
\begin{tabular}{@{}l@{}rrrrrrrrc@{}}
\toprule
Library & Contributors & Forks & Watchers & Stars & Issues & PRs &Coverage & License \\ 
\midrule
TorchGeo & 108 & 473 & 56 & 3,663 & 173 & 2,339 & 100\% & MIT \\
OTB & 41 & 121 & 40 & 375 & 3 & 24 & 56\% & Apache-2.0 \\
TerraTorch & 40 & 101 & 23 & 601 & 92 & 634 & 55\% & Apache-2.0 \\
DeepForest & 35 & 217 & 17 & 657 & 96 & 530 & 86\% & MIT \\
Raster Vision & 33 & 393 & 71 & 2,163 & 40 & 1,465 & 90\% & Apache-2.0 \\
samgeo & 24 & 351 & 59 & 3,408 & 33 & 154 & 13\% & MIT \\
spopt & 23 & 55 & 12 & 341 & 30 & 265 & 77\% & BSD-3-Clause \\
SITS & 17 & 86 & 27 & 515 & 26 & 653 & 91\% & GPL-2.0 \\
SPy & 15 & 146 & 35 & 636 & 25 & 38 & 69\% & MIT \\
srai & 15 & 27 & 12 & 316 & 104 & 280 & 92\% & Apache-2.0 \\
ML4Floods & 13 & 42 & 18 & 170 & 1 & 73 & 0\% & LGPL-3.0 \\
GEO-Bench & 13 & 12 & 12 & 153 & 8 & 9 & 51\% & Apache-2.0 \\
scikit-eo & 8 & 24 & 6 & 206 & 4 & 14 & 32\% & Apache-2.0 \\
GeoAI & 7 & 202 & 33 & 1,587 & 24 & 174 & 6\% & MIT \\
Myria3D & 7 & 31 & 14 & 259 & 18 & 79 & 57\% & BSD-3-Clause \\
GeoTessera & 6 & 14 & 4 & 146 & 11 & 9 & 15\% & ISC \\
OTBTF & 5 & 39 & 11 & 166 & 22 & 19 & 55\% & Apache-2.0 \\
GeoDeep & 3 & 24 & 9 & 326 & 3 & 3 & 0\% & AGPL-3.0 \\
\bottomrule
\end{tabular}
}
\end{table}

\begin{table}[tb]
\centering
\caption{Download statistics for popular GeoML libraries (as of September 2025). Note that weekly download metrics may be volatile. OTB and OTBTF are not distributed on any package index and therefore download statistics are not available. SITS includes both CRAN and PyPI downloads.}
\label{tab:downloads}
\begin{tabular}{lrrrrr}
\toprule
 & \multicolumn{3}{c}{PyPI/CRAN} & Conda & Total \\
\cmidrule(lr){2-4}
Library & Last Week & Last Month & All Time & All Time & All Time \\ 
\midrule
TorchGeo & 69,430 & 200,906 & 845,353 & 40,949 & 886,302 \\
OTB & 0 & 0 & 0 & 0 & 0 \\
TerraTorch & 3,870 & 29,514 & 130,473 & 0 & 130,473 \\
DeepForest & 2,483 & 6,526 & 988,841 & 97,503 & 1,086,344 \\
Raster Vision & 4,666 & 8,613 & 221,409 & 5,872 & 227,281 \\
samgeo & 2,319 & 6,692 & 274,609 & 54,307 & 328,916 \\
spopt & 13,289 & 44,498 & 1,122,138 & 251,912 & 1,374,050 \\
SITS & 477 & 1,967 & 22,502 & 136,850 & 159,352 \\
SPy & 17,720 & 59,792 & 1,445,738 & 150,137 & 1,595,875 \\
srai & 661 & 1,962 & 51,103 & 0 & 51,103 \\
ML4Floods & 90 & 195 & 10,254 & 0 & 10,254 \\
GEO-Bench & 609 & 1,818 & 58,756 & 0 & 58,756 \\
scikit-eo & 515 & 598 & 22,281 & 0 & 22,281 \\
GeoAI & 1,765 & 7,277 & 63,399 & 16,747 & 80,146 \\
Myria3D & 13 & 33 & 4,429 & 0 & 4,429 \\
GeoTessera & 348 & 1,218 & 3,725 & 0 & 3,725 \\
OTBTF & 0 & 0 & 0 & 0 & 0 \\
GeoDeep & 229 & 920 & 18,540 & 0 & 18,540 \\
\bottomrule
\end{tabular}
\end{table}

As of September 2025, the current set of popular GeoML libraries under active development include: TorchGeo, OTB, TerraTorch, DeepForest, Raster Vision, samgeo, spopt, SITS, SPy, srai, ML4Floods, GEO-Bench, scikit-eo, GeoAI, Myria3D, GeoTessera, OTBTF, and GeoDeep. Table~\ref{tab:features} summarizes the machine learning backend (e.g., PyTorch, TensorFlow, scikit-learn), dataset/model availability, and any notable features such as a command-line interfaces (CLI), graphical user interfaces (GUI), support for automatic reprojection and resampling, SpatioTemporal Asset Catalogs (STAC), or time-series analysis for each of these libraries. Tables~\ref{tab:github} and \ref{tab:downloads} display a number of metrics useful for gauging the popularity and stability of each library.\footnote{All tables are maintained in perpetuity at \url{https://torchgeo.rtfd.io/en/latest/user/alternatives.html}. If any metrics are out of date, please open a pull request to update them.} While the number of GitHub stars and PyPI downloads are useful for gauging popularity, the number of contributors and test coverage are much more reliable for gauging the stability and long-term success of the library. Finally we describe the abstractions in three popular libraries: TorchGeo, eo-learn, and Raster Vision. We also assess the typical user base, limitations, and scope of each library.

\subsubsection{TorchGeo}

\textbf{TorchGeo} is a PyTorch domain library designed to unite machine learning and remote sensing experts under a single platform. It defines two broad classes of datasets: \texttt{GeoDataset} for uncurated raster and vector data files and \texttt{NonGeoDataset} for curated benchmark datasets. 

Like other libraries on this list, users can directly use \texttt{GeoDataset} and its subclasses, \texttt{RasterDataset} and \texttt{VectorDataset}, to load and process arbitrary raster and vector data. Each dataset consists of a geopandas \texttt{GeoDataFrame} index, allowing different datasets to be intelligently composed based on spatiotemporal intersection or union. The resulting dataset can then be queried by a spatiotemporal slice, allowing users to quickly load small patches of imagery and masks from large collections of raster scenes. Reprojection and resampling are automatically handled by rasterio and fiona, and users can specify which spectral bands they want to load. To ease spatiotemporal indexing, TorchGeo defines a \texttt{GeoSampler} interface, allowing users to select various random (for training) or sequential (for inference) sampling strategies.

TorchGeo also includes a number of curated ML-ready benchmark datasets designed to train and evaluate models for specialized tasks such as crop type mapping, biomass estimation, or disaster response. Each of these datasets provides an input image and an output target, such as a semantic segmentation mask or object detection bounding box. In total, TorchGeo provides over 125 built-in geospatial and benchmark datasets, more than all other libraries combined.

Another primary focus of TorchGeo is on providing pretrained foundation models capable of being finetuned for any task. TorchGeo provides over 120 such model weights, including models pretrained on Landsat, NAIP, and Sentinel imagery and newer ``sensor-agnostic'' foundation models like Scale-MAE \citep{reed2023scale}, DOFA \citep{xiong2024neural}, CROMA \citep{fuller2023croma}, Panopticon \citep{waldmann2025panopticon}, and Copernicus-FM \citep{wang2025towards}.

TorchGeo places a great deal of emphasis on documentation and testing, with 100\% test coverage on three Python versions, three platforms, and the minimum and maximum supported dependency versions. It also has more than double the number of contributors as any other library on this list, demonstrating its widespread adoption by the community. Its component-based nature, providing PyTorch-compatible datasets and models for the community, naturally allows other libraries like TerraTorch and GeoAI to build on top of TorchGeo, extending its features and capabilities.

\subsubsection{eo-learn}

\textbf{eo-learn} is a Python library that aims to bridge the gap between the remote sensing and data science ecosystems. It is highly coupled with the commercial Sentinel Hub library and ecosystem, and enables users to build reusable workflows based around the \texttt{EOPatch}, \texttt{EOTask}, and \texttt{EOWorkflow} abstractions. An \texttt{EOPatch} contains all types of data for a given bounding box location --- raster data with time series support, vector data, and one-off masks like land cover information. \texttt{EOTask}s operate on \texttt{EOPatch}es and transform them, for example performing cloud masking, computing spectral indices, classical feature extraction pipelines, and/or classification. Finally, \texttt{EOTask}s are organized into computational graphs that exchange patch objects and run in \texttt{EOWorkflow}s, allowing for parallelization and record keeping.

For data I/O, eo-learn relies on GDAL (typically through rasterio) for reading geospatial raster data, and also uses OpenCV, pandas, SciPy, and Zarr under the hood. This means it can handle common raster formats and even work with large out-of-core datasets (Zarr provides chunked, on-disk array storage). For vector data and geometric operations, eo-learn uses GeoPandas, enabling integration of shapefiles or GeoJSON data for tasks like sampling points or computing zonal statistics. NumPy is used as the primary foundation for array transformations and manipulations within tasks.

Time-series modeling is not explicitly provided as a specialized feature, although eo-learn can certainly process time-indexed imagery (e.g., stacking time frames in an \texttt{EOPatch} object) --- but it lacks specialized time-series ML models or temporal analysis tools out-of-the-box. eo-learn is best suited for Earth observation practitioners and data scientists who need to preprocess satellite imagery and derive features for modeling. Its ease of integration with scikit-learn makes it ideal for prototyping land cover classifications or regressions on remote sensing data when deep learning is not required.

\subsubsection{Raster Vision}

\textbf{Raster Vision} is a Python library for training and deploying deep learning models with satellite imagery. It was originally developed by Azavea and has been maintained by Element 84 since February of 2023. Similar to eo-learn, Raster Vision is built around the idea of data pipelines. The input to a Raster Vision pipeline is imagery and labels over given areas of interest (AOIs) while the output is a trained model formatted in a ``bundle'' for deployment. The pipeline steps include:
\begin{enumerate}
    \item \textbf{Analyze} - compute dataset-level statistics
    \item \textbf{Chip} - convert geospatial data into uniformly sized patches that can be input into a model in a batched manner
    \item \textbf{Train} - train a machine learning model via a \texttt{Learner} abstraction (typically using a PyTorch backend)
    \item \textbf{Predict and Evaluate} - run inference with a trained model over validation and test data and compute performance metrics
    \item \textbf{Bundle} - Create a model bundle to be used in deployments
\end{enumerate}

Raster Vision creates abstractions for every step in the above pipeline. For example, input data is represented by source classes: \texttt{RasterSource}, with subclasses such as \texttt{RasterioSource} or \texttt{XarraySource}, \texttt{VectorSource}, and \texttt{LabelSource}. Different sources are combined into \texttt{Scene}s --- collections of data over the same area of interest --- and finally a PyTorch DataLoader-compatible \texttt{GeoDataset} object that can be used with various training pipelines (e.g., PyTorch Lightning). 

\begin{figure}[htb]
    \centering
    \includegraphics[width=0.7\linewidth]{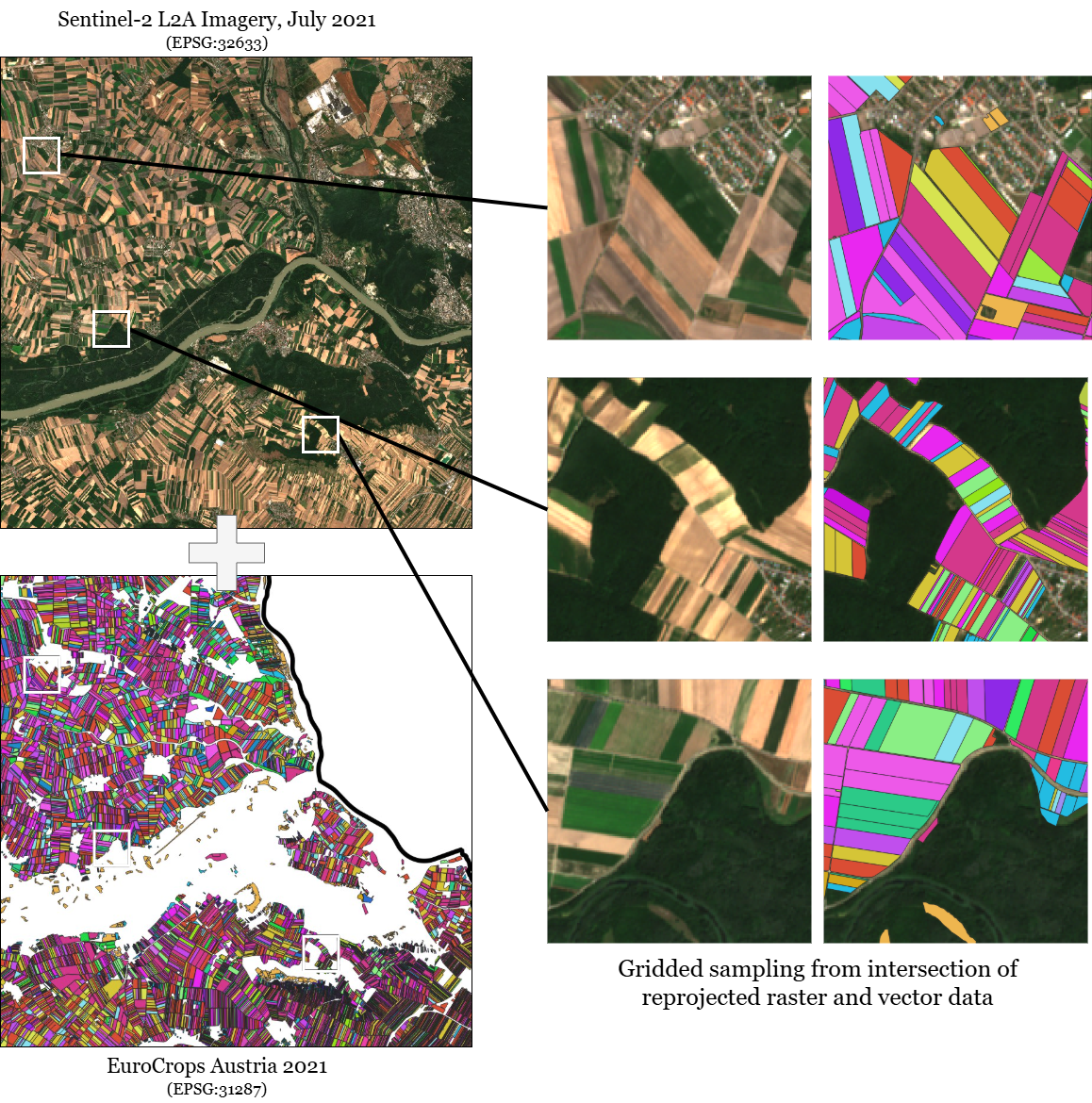}
    \caption{A common use case for GeoML practitioners is joining raster data, such as satellite imagery (\textbf{top left}), with vector data, such as crop masks (\textbf{bottom left}), then randomly sampling patches from the intersection of these datasets to use in modeling pipelines (\textbf{right}).}
    \label{fig:eurocrop_example}
\end{figure}

\section{Application Example}\label{sec:Application}

In order to demonstrate the power and potential of GeoML libraries, let us consider the following example. A researcher has access to hundreds (or even thousands) of raster images and one or more vector files containing polygon labels. They would like to train a model to perform semantic segmentation, assigning a class label to every pixel in the image. %
Each raster and vector file in the dataset may have a different CRS or resolution, and may or may not have spatiotemporal overlap. This example is representative of a broad class of tasks in remote sensing, and can apply to anything from building or road segmentation, to land use/land cover mapping, to deforestation and natural disaster analysis.

To make this more concrete, let us say that the raster images are from Sentinel-2 \citep{Drusch2012Sentinel2} and the vector files are from EuroCrops \citep{schneider2023eurocrops} (see Figure~\ref{fig:eurocrop_example}). The goal is to perform crop type mapping, with each pixel being assigned to one of \(\sim\)200 crop classes. This requires reprojection, rasterization, and chipping of each image and vector file into smaller image patches that can be passed through the model. For the model, we will use a simple U-Net \citep{ronneberger2015u} architecture with a ResNet-50 \citep{he2016deep} backbone.

Now, let us see how this task can be solved with various levels of abstraction using TorchGeo. TorchGeo is available on PyPI and Conda Forge, and can easily be installed using:
\begin{minted}{console}
> pip install torchgeo
\end{minted}
TorchGeo is designed with the same API as other PyTorch domain libraries. The below code should look familiar to anyone with experience using libraries like torchvision and PyTorch Lightning. The following examples demonstrate three different levels of abstraction, each with their own pros and cons.\footnote{Maintained and well-tested notebooks for many of these application examples can be found at: \url{https://torchgeo.rtfd.io/en/latest/tutorials/getting_started.html}.}

\subsection{Pure PyTorch}

For maximum control and customization, TorchGeo datasets, samplers, and pre-trained models can be used directly. While the data loader setup is a bit custom, the rest of the training and evaluation pipeline is identical to other PyTorch domain libraries.

First, we import everything we will later require.
\begin{minted}{python}
import kornia.augmentation as K
import segmentation_models_pytorch as smp
import torch
from torch import nn, optim
from torch.utils.data import DataLoader
from torchgeo.datasets import (
    EuroCrops, Sentinel2, random_grid_cell_assignment
)
from torchgeo.models import ResNet50_Weights
from torchgeo.samplers import (
    GridGeoSampler, RandomGeoSampler
)
\end{minted}
Given a user-defined local directory or list of remote file paths, we next instantiate PyTorch \texttt{Dataset} objects for Sentinel-2 and EuroCrops data. Given a spatiotemporal slice, this dataset is responsible for loading the data at that time and location from disk. We then compute the spatiotemporal intersection of these datasets. All data is automatically reprojected to a shared CRS and rasterized on the fly.
\begin{minted}{python}
sentinel2 = Sentinel2(paths='...')
eurocrops = EuroCrops(paths='...', download=True)
dataset = sentinel2 & eurocrops
\end{minted}
Next, we overlay a \(10 \times 10\) grid over the dataset. Each grid cell is randomly assigned to a train, validation, or test split using an 80:10:10 split. Note that TorchGeo has other geospatial splitting strategies that may be more suitable for out-of-distribution evaluation.
\begin{minted}{python}
train_ds, val_ds, test_ds = random_grid_cell_assignment(
    dataset, [0.8, 0.1, 0.1], grid_size=10
)
\end{minted}
Next, we instantiate PyTorch \texttt{Sampler} objects for each split. The sampler is responsible for telling the dataset where and when to load data from. Here, we use random sampling at training time to maximize the diversity of the training data and grid-based sampling at evaluation time to avoid overlap or missing locations.
\begin{minted}{python}
train_sr = RandomGeoSampler(train_ds, size=224)
val_sr = GridGeoSampler(val_ds, size=224, stride=224)
test_sr = GridGeoSampler(test_ds, size=224, stride=224)
\end{minted}
All TorchGeo datasets and samplers are compatible with PyTorch's built-in \texttt{DataLoader} class. Here, we set the batch size and tell the data loader where it is allowed to sample from.
\begin{minted}{python}
batch_size = 64
train_dl = DataLoader(train_ds, batch_size, sampler=train_sr)
val_dl = DataLoader(val_ds, batch_size, sampler=val_sr)
test_dl = DataLoader(test_ds, batch_size, sampler=test_sr)
\end{minted}
We use Kornia for our preprocessing and data augmentation transforms. Preprocessing is required to ensure that all inputs are in the 0--1 range, regardless of whether the original imagery was in uint8 or uint12 format. Data augmentation is helpful to artificially inflate the size of the training dataset.
\begin{minted}{python}
preprocess = K.Normalize(0, 10000)
augment = K.ImageSequential(
    K.RandomHorizontalFlip(), K.RandomVerticalFlip()
)
\end{minted}
Next, we instantiate our model. TorchGeo comes with 120+ model weights pre-trained on SAR, RGB, MSI, and HSI remote sensing imagery. In this case, we use a model pre-trained on all 13 bands of Sentinel-2 imagery using self-supervised learning \citep{wang2023ssl4eo}.
\begin{minted}{python}
model = smp.Unet('resnet50', in_channels=13, classes=200)
weights = ResNet50_Weights.SENTINEL2_ALL_DINO
model.encoder.load_state_dict(weights.get_state_dict())
\end{minted}
We next transfer our model to the GPU. Note that this step may look different for NVIDIA, AMD, and Apple Silicon GPUs.
\begin{minted}{python}
device = 'cuda' if torch.cuda.is_available() else 'cpu')
model = model.to(device)
\end{minted}
For our loss function, we choose cross-entropy loss, which works well for multiclass classification and semantic segmentation tasks. For our optimizer, we choose stochastic gradient descent (SGD). Users can and should experiment with fancier optimizers here.
\begin{minted}{python}
loss_fn = nn.CrossEntropyLoss()
optimizer = optim.SGD(model.parameters(), lr=1e-2)
\end{minted}
We define a train function containing the forward and backward pass of our model for a single epoch.
\begin{minted}{python}
def train(dataloader):
    model.train()
    total_loss = 0
    for batch in dataloader:
        x = batch['image'].to(device)
        y = batch['mask'].to(device)
        x = preprocess(x)
        x = augment(x)

        # Forward pass
        y_hat = model(x)
        loss = loss_fn(y_hat, y)
        total_loss += loss.item()

        # Backward pass
        loss.backward()
        optimizer.step()
        optimizer.zero_grad()

    print(f'Loss: {total_loss:.2f}')
\end{minted}
We define a similar evaluation function containing the forward pass of our model and compute the overall accuracy of a single epoch.
\begin{minted}{python}
def evaluate(dataloader):
    model.eval()
    correct = 0
    with torch.no_grad():
        for batch in dataloader:
            x = batch['image'].to(device)
            y = batch['label'].to(device)
            x = preprocess(x)

            # Forward pass
            y_hat = model(x)
            y_hat_hard = y_hat.argmax(1) == y
            correct += y_hat_hard.sum().item()

    correct /= len(dataloader.dataset)
    print(f'Accuracy: {correct:.0%
\end{minted}
Finally, we combine everything and train and evaluate our model over 100 epochs. Validation accuracy can be used for early stopping and for hyperparameter tuning. Once an optimal set of hyperparameters is found, we report the final test accuracy.
\begin{minted}{python}
for epoch in range(100):
    print(f'Epoch: {epoch}')
    train(train_dataloader)
    evaluate(val_dataloader)

evaluate(test_dataloader)
\end{minted}

\subsection{PyTorch Lightning}

While the above code has the most flexibility and can be easily customized to add preprocessing or data augmentation, it can be daunting for many users. Beginners are likely to make mistakes when computing accuracy metrics, while experts may require a lot of code duplication if they want to, for example, evaluate several model architectures on dozens of different datasets. To simplify the training and evaluation pipeline, TorchGeo offers PyTorch Lightning \citep{lightning} integration. PyTorch Lightning introduces two ideas: \textit{data modules}, collections of train/val/test datasets and the transforms applied to them, and \textit{modules}, reusable recipes for tasks like classification, regression, or semantic segmentation. TorchGeo provides built-in data modules for many datasets and modules for:
\begin{itemize}
    \item Classification (binary, multiclass, multilabel)
    \item Regression (imagewise, pixelwise)
    \item Semantic Segmentation (binary, multiclass, multilabel)
    \item Change Detection (binary, multiclass, multilabel)
    \item Object Detection
    \item Instance Segmentation
    \item Self-Supervised Learning (BYOL \citep{grill2020bootstrap}, MoCo \citep{he2020momentum}, SimCLR \citep{chen2020simple})
\end{itemize}
The following code can be easily extended to support TensorBoard logging \citep{tensorboard}, early stopping, and model checkpointing, and implements all of the steps from the previous section with significantly fewer lines of code.
\begin{minted}{python}
from lightning.pytorch import Trainer
from torchgeo.datamodules import Sentinel2EuroCropsDataModule
from torchgeo.models import ResNet50_Weights
from torchgeo.trainers import SemanticSegmentationTask

model = SemanticSegmentationTask(
    model='unet',
    backbone='resnet50',
    weights=ResNet50_Weights.SENTINEL2_ALL_DINO,
    in_channels=13,
    num_classes=200,
    lr=1e-2,
)
datamodule = Sentinel2EuroCropsDataModule(
    sentinel2_paths='...',
    eurocrops_paths='...',
    batch_size=64,
    patch_size=224,
)
trainer = Trainer(max_epochs=100)
trainer.fit(model=model, datamodule=datamodule)
trainer.test(model=model, datamodule=datamodule)
\end{minted}

\subsection{Command Line}

In the research community, easy experimentation and reproducibility are paramount. TorchGeo offers a command-line interface based on LightningCLI, with support for command-line, YAML, and JSON configuration. The following YAML file can be used to reproduce an experiment from a paper without writing a single line of code.
\begin{minted}{yaml}
model:
  class_path: SemanticSegmentationTask
  init_args:
    model: 'unet'
    backbone: 'resnet50'
    weights: 'ResNet50_Weights.SENTINEL2_ALL_DINO'
    in_channels: 13
    num_classes: 200
    lr: 1e-2
data:
  class_path: Sentinel2EuroCropsDataModule
  init_args:
    batch_size: 64
    patch_size: 224
  dict_kwargs:
    sentinel2_path: '...'
    eurocrops_path: '...'
\end{minted}
Once TorchGeo is installed, the user simply needs to run commands like \texttt{fit} or \texttt{test} to train and evaluate their model.
\begin{minted}{console}
> torchgeo fit --config config.yaml
> torchgeo test --config config.yaml --ckpt_path=...
\end{minted}

\section{Best Practices and Open Issues}\label{sec:Guides}

With so many successful or abandoned GeoML libraries over the years, there is a wealth of information on best practices to follow (and also what to avoid). Despite significant progress towards forming a consensus on code and data hosting repositories, there are still many open issues for future generations to tackle.

\subsection{File Formats}

One of the primary purposes of GeoML libraries is to load and preprocess data. Thus, choosing the right file format is critical to achieve efficient I/O and keep the GPU busy. While GeoML libraries themselves are often agnostic to file format and support a wide range of file types, \textit{users} of GeoML libraries can drastically improve the speed of training and inference by choosing the right file type.

For uncurated raster and vector data layers, GeoTIFF and ESRI Shapefile have long been standard. In particular, so-called ``cloud optimized GeoTIFFs'' (COGs) support fast and efficient windowed reading, allowing tools like TorchGeo to load small image patches without reading the entire file from disk. Preprocessing, including manual reprojection, target aligned pixels (TAP), and compression, can all improve I/O performance \citep{torchgeo}.

For curated ML-ready benchmark datasets, geospatial metadata is often unnecessary or absent. Typical file formats include PNG and JPEG, with newer file formats like Parquet and Zarr becoming more common for larger datasets. Language- or software-specific file formats like PyTorch's .pth, NumPy's .npy, or Python's pickle should be avoided to ensure cross-language support.

Large-scale datasets commonly used for self-supervised learning and foundation model pretraining require their own careful consideration. Datasets such as SSL4EO \citep{wang2023ssl4eo,stewart2023ssl4eo}, SatlasPretrain \citep{bastani2023satlaspretrain}, and CopernicusPretrain \citep{wang2025towards} consist of anywhere from millions to billions of files, easily enough to overwhelm all but the largest compute clusters. Modern formats like WebDataset \citep{webdataset} and LitData \citep{litdata} offer support for sharding and streaming data, avoiding inode and storage limits.

\subsection{Distribution}

GitHub is the primary hosting platform for GeoML libraries, with every single library on the list being developed or mirrored on GitHub. While many geospatial software packages are still hosted on private servers like OSGeo and bug reports are filed through TRAC, GitHub has long been popular within the ML community. Having a single shared platform for everything from version control to continuous integration to bug reporting makes it easier for potential users to find your software and potential contributors to improve your software.

While GitHub can be used for release distribution, the vast majority of successful GeoML libraries are distributed through language-specific servers like PyPI (Python), CRAN (R), or Conda-Forge (Python/R). In fact, most of the abandoned GeoML libraries on our list either never had stable releases, or were only distributed through \texttt{git clone} or \texttt{docker pull}. In the Python ecosystem, pre-compiled binaries (bdists) such as wheels are critical for ease of installation, especially on Windows.

GeoML datasets and pre-trained model weights can be found on a variety of platforms, with personal Google Drive, OneDrive, and Baidu Drive being common. However, these platforms should be avoided for serious scientific advances, as it is too easy to accidentally delete the wrong file or directory. Instead, dedicated data hosting platforms like Zenodo or Hugging Face should be used instead. While AWS S3 buckets and Source Cooperative repositories are useful for extremely large datasets, they lack stable download URLs and checksumming capabilities required to ensure reproducibility.

\subsection{Licensing}

One of the most important considerations for developers and users of GeoML software is the license under which it is distributed. Often overlooked by researchers, the license allows the owner of the IP (copyright, trademark, and/or patent) to explicitly grant certain \textit{permissions}, with certain \textit{limitations}, subject to certain \textit{conditions}. For example, the popular MIT license grants the user \textit{permission} to redistribute and modify the software, including commercial and private use, \textit{limiting} liability and warranty of the developer, under the \textit{condition} that the user preserves the license and copyright notice. In this sense, licenses protect both the users and contributors.

There are two broad families of licenses: permissive and copyleft. Permissive licenses like MIT and Apache-2.0 are generally more relaxed, with the only \textit{condition} that the license must be preserved. Copyleft licenses like the GNU Public License (GPL), on the other hand, impose strict constraints on users such that any code bundled with the software must be released under the same license. Copyleft licenses are generally incompatible with industry, limiting adoption by many communities. With the exception of SITS, ML4Floods, and GeoDeep, every single GeoML library under active development is released under a permissive license (see Table~\ref{tab:github}), reflecting the broader Python and ML communities. SITS, like R itself, is licensed under GPL, restricting its use primarily to academic settings.

While software licenses can technically be used for data and models, many of the clauses in these licenses are not applicable to data or models. Instead, the Creative Commons family of licenses is more common, including CC0 (public domain), CC-BY (Attribution), CC-BY-SA (Attribution-ShareAlike), and CC-BY-NC (Attribution-NonCommercial). Responsible AI Licenses (RAIL) are another family of licenses tailored specifically for the AI community, with subcategories for Data, Apps, Models, and Source code. However, it is worth noting that CC-BY-NC and OpenRAIL are not considered to be open source licenses, as they restrict certain usage.

The important thing for both code and data is to choose a license\footnote{\url{https://choosealicense.com/}} and stick with it. In particular, datasets without licenses are unfortunately pervasive and challenging for GeoML libraries. While rarely enforced, it is technically illegal to even download let alone use or redistribute data without a license.

\subsection{Continuous Integration}

A good software library is not complete without \textit{continuous integration} (CI), suites of checks and tests to ensure software quality that typically run on every commit, branch, pull request, and release. GitHub Actions is currently the dominant CI platform for free, public libraries, replacing Travis CI and CircleCI which dominated before its release. 

CI can and should consist of a variety of different types of testing. The most obvious is unit testing, designed to ensure that individual functions and classes behave as they are designed to. Testing frameworks, including pytest for Python and testthat for R, handle the heavy lifting, allowing one to quickly add test cases (expected output given a certain input). Many libraries also run integration tests on release branches, ensuring that functions and classes can be integrated into a larger framework. Test coverage is particularly important as it ensures that the majority of the library is touched by unit tests. Test coverage can be reported by services including Codecov and Coveralls. Software libraries with less than 80\% test coverage are generally not recommended for production use and should be considered unstable.

Even more important than unit testing is documentation. Tools like Sphinx can be used to build and test the documentation, and documentation hosting sites like ReadTheDocs provide CI support, allowing developers to see the documentation generated by each pull request. Sphinx can also automatically generate API documentation from function and class docstrings, allowing you to easily maintain high-quality documentation.

Dynamically typed languages like Python can also benefit from type hints. Tools like mypy, pytype, pyright, pyre, and ty can be used to enforce strict static typing, preventing a number of bugs caused by bad typing practices. Tools like Sphinx can even use these type hints when generating API documentation. Similarly, tools like flake8, isort, black, and ruff can be used to enforce style guides, resulting in well-formatted code. While these may not seem important for single-author projects, they become critical for maintaining projects with hundreds of contributors.

\subsection{Open Issues}

Testing and reproducibility \citep{heil2021reproducibility} remain the biggest challenges faced by GeoML libraries at the moment. While many actively maintained GeoML libraries have over 90\% test coverage (see Table~\ref{tab:github}), the majority fall far below, with some having no testing at all. A lack of tests puts these projects at risk of both bugs (unexpected or incorrect behavior) and regressions (when a fixed bug resurfaces later). In addition, tests can ensure software stability, warning developers of unintentional backwards-incompatible changes and allowing them to mitigate or document the change before making a release.

Test coverage itself is not the only important metric. Just because a line of code can be executed without raising an error does not mean it produces a desirable result. Machine learning is inherently stochastic, with non-determinism originating from the data (cross validation split, data augmentation, random shuffling), the model (initial weights, dropout layers), and the optimization process (SGD, momentum, learning rate scheduler). While setting a random seed can alleviate some of these sources of non-determinism, reproducibility is not guaranteed across software versions, platforms, and accelerators (CPU, GPU, TPU). This makes testing challenging for developers and perfect reproducibility practically impossible for users.

As datasets and models continue to grow in size, computational performance becomes increasingly important. Maximizing I/O and data transfer speeds, especially for parallel filesystems and on distributed compute, remains an important challenge. Libraries like Kornia \citep{eriba2019kornia} address some of these challenges by implementing data augmentations directly in PyTorch or Rust, providing significant speedups over pure Python implementations. It may be possible to achieve similar speedups for reprojection and resampling, often the biggest bottlenecks for geospatial data processing, by implementing these operations in PyTorch or CUDA and performing them on the GPU.

\section{Future Implications}\label{sec:Outlook}

With increased interest in and funding for geospatial intelligence by industry, the future of GeoML libraries looks promising. Below, we list our own predictions for what the next five years will look like.

\subsection{Foundation Models to Embeddings}

Foundation models \citep{bommasani2021opportunities} --- large, task-agnostic deep learning models pretrained on massive amounts of data --- have dominated the last few years of research (see the self-supervised learning chapter of this book). We expect this dominance to continue, especially in the geospatial domain where scientists so often deal with limited labeled data. While most current research focuses on EO foundation models for SAR and MSI satellite imagery, we expect foundation models to expand to a greater number of modalities, including HSI, UAV, and LiDAR data. 

We also anticipate the development of unified foundation models for the land surface, ocean, and atmosphere \citep{zhu2024foundations}. This is especially important in the area of weather and climate modeling, where these processes are all interconnected and cannot be individually modeled. Application domains like air pollution and nowcasting provide new challenges for GeoML libraries, forcing the incorporation of real-time point data in unstructured graphs instead of curated raster data.

The recent release of a number of high-profile foundation model embeddings, including Major TOM \citep{czerkawski2024global} and AlphaEarth Foundations \citep{brown2025alphaearth}, raises the question: are GeoML libraries still necessary? Deep learning can be very computationally demanding and requires technical expertise that many users lack, while pre-computed global embeddings allow users to run simple linear probing layers or $k$-NN models directly on foundation model outputs. However, GeoML libraries are still necessary to generate new embeddings, and foundation model fine-tuning still offers higher performance. Conversely, embeddings have opened the door for a new type of GeoML library, like GeoTessera, which ease the use of large-scale embeddings.

\subsection{Growing Focus on Reuse and Ease of Use}

The majority of the history of GeoML libraries has seen a ``reinvention of the wheel'', with new libraries introducing clever ideas and solutions for common challenges without reusing older libraries. This is especially true for pipeline-based libraries like Raster Vision and eo-learn, which are easy to use but difficult to build on top of. The introduction of component-based libraries like TorchGeo has changed this, with libraries like GEO-Bench, TerraTorch, and GeoAI supplementing instead of replacing TorchGeo.

Foundation models and embeddings have shown us that ease of use is often more important than accuracy. We expect further abstractions on top of component-based libraries providing low-code or no-code solutions for GIS domain experts. A handful of machine learning plugins already exist for ArcGIS and QGIS, but lack the power and features of existing GeoML libraries. Vision--language foundation models like DOFA-CLIP \citep{xiong2025dofaclip} present an interesting opportunity for zero-shot learning, in which the user can ask the model to perform a task using natural language.

\subsection{Independent Governance and Software Foundations}

However, successful software is often made by multiple people working together on behalf of a company or research institution. This incubation process provides the necessary funding and advertising needed for the project to take off. After gaining popularity, successful software can quickly outgrow the organization that created it. Once this happens, independent governance is required to allow the project to continue to grow. Examples of this include PyTorch (developed by Meta) and TensorFlow/Keras/JAX (developed by Google), which are now governed by independent teams of maintainers.

Several GeoML libraries have graduated beyond this incubation status and achieved open and independent governance. OTB gained independence from CNES and announced its new open governance model in March of 2015. PySAL has always been independent and adopted a formal governance structure in February of 2019. TorchGeo gained independence from Microsoft in August of 2025 and now holds monthly Technical Steering Committee meetings open to the public. All three of these libraries have cultivated successful open source ecosystems, with further GeoML libraries extending their capabilities: OTB (OTBTF), PySAL (spopt), and TorchGeo (GEO-Bench, GeoAI, TerraTorch).

An interesting question is what role software foundations should play in this incubation process. Many geospatial and ML software foundations exist, including the Open Source Geospatial Foundation (OSGeo), the PyTorch Foundation, the Linux Foundation AI \& Data, and the AI Alliance. OTB and TorchGeo already belong to OSGeo, while TerraTorch recently joined the AI Alliance. These foundations provide advertising, financial support, and legal support for a variety of software projects, promoting collaboration between member projects. These software foundations could provide a neutral home for many such GeoML libraries.

\section*{Acknowledgments}

The authors would like to thank the contributors to and maintainers of all GeoML software for their tireless efforts in building much-needed software infrastructure for this domain. In particular, we would like to thank Nicolas Audebert (author of DeepNetsForEO and DeepHyperX) and Gilberto Camara (author of SITS) for sharing their knowledge about the early history of GeoML software. We would also like to thank Robin Cole and Eduardo Lacerda for maintaining listings of all major GeoML libraries throughout recent years. The work was supported in part by the National Science Foundation (NSF) through awards IIS 21-31335, OAC 21-30835, DBI 20-21898, as well as a C3.ai research award.

\printbibliography

\end{document}